\documentclass{article}

\usepackage{hyperref}       
\usepackage{url}            
\usepackage{booktabs}       
\usepackage{amsfonts}       
\usepackage{nicefrac}       
\usepackage{microtype}      
\usepackage{graphicx}
\usepackage{amsmath,amssymb} 
\usepackage{color}
\usepackage{epsfig}
\usepackage{tabularx}
\usepackage{booktabs}
\usepackage{multirow}
\usepackage{array}
\usepackage{xspace}
\usepackage{balance}
\usepackage{enumitem}
\usepackage{mathtools}
\usepackage{rotating}
\usepackage{tabulary}
\usepackage[table,dvipsnames]{xcolor}
\usepackage{subcaption}
\usepackage[export]{adjustbox}
\usepackage{ctable}
\usepackage{diagbox}
\usepackage{makecell}
\usepackage{bbding}
\usepackage{pifont}
\usepackage{stfloats}

\newcommand{\cmark}{\text{\ding{51}}}
\newcommand{\xmark}{\text{\ding{55}}}

\newcommand{\eg}{\textit{e}.\textit{g}.}

\newcolumntype{C}[1]{>{\centering\let\newline\\\arraybackslash\hspace{0pt}}m{#1}}

\def\name{BLIP}

\usepackage[accepted]{icml2022}

\icmltitlerunning{BLIP: Bootstrapping Language-Image Pre-training for Unified Vision-Language Understanding and Generation}

\begin{document}

\twocolumn[
\icmltitle{BLIP: Bootstrapping Language-Image Pre-training for \\ Unified Vision-Language Understanding and Generation}



\icmlsetsymbol{equal}{*}

\begin{icmlauthorlist}
\icmlauthor{Junnan Li}{}
\icmlauthor{Dongxu Li}{}
\icmlauthor{Caiming Xiong}{}
\icmlauthor{Steven Hoi}{}
\\
\icmlauthor{Salesforce Research}{}
\\
\url{https://github.com/salesforce/BLIP}{}
\end{icmlauthorlist}

\icmlcorrespondingauthor{Junnan Li}{junnan.li@salesforce.com}

\icmlkeywords{vision-language}

\vskip 0.3in
]




\begin{abstract}
 Vision-Language Pre-training (VLP) has advanced the performance for many vision-language tasks.
 However,
 most existing pre-trained models only excel in either understanding-based tasks or generation-based tasks.
 Furthermore,
 performance improvement has been largely achieved by scaling up the dataset with noisy image-text pairs collected from the web,
 which is a suboptimal source of supervision.
 In this paper,
 we propose \name,
 a new VLP framework which transfers flexibly to both vision-language understanding and generation tasks.
 \name~effectively utilizes the noisy web data by bootstrapping the captions, where a captioner generates synthetic captions and a filter removes the noisy ones.
 We achieve state-of-the-art results on a wide range of vision-language tasks,
 such as image-text retrieval (+2.7\% in average recall@1), image captioning (+2.8\% in CIDEr),
 and VQA (+1.6\% in VQA score).
 \name~also demonstrates strong generalization ability when directly transferred to video-language tasks in a zero-shot manner.
 Code, models, and datasets are released.

\end{abstract} 
\vspace{-3ex}
\section{Introduction}
Vision-language pre-training has recently received tremendous success on various multimodal downstream tasks.
However, existing methods have two major limitations:

(1) Model perspective: most methods either adopt an encoder-based model~\cite{clip,ALBEF}, or an encoder-decoder~\cite{VL_T5,simvlm} model.
However, encoder-based models are less straightforward to directly transfer to text generation tasks (\eg~image captioning), whereas encoder-decoder models have not been successfully adopted for image-text retrieval tasks.

(2) Data perspective: most state-of-the-art methods (\eg, CLIP~\cite{clip}, ALBEF~\cite{ALBEF}, SimVLM~\cite{simvlm}) pre-train on image-text pairs collected from the web. 
Despite the performance gain obtained by scaling up the dataset,
our paper shows that the noisy web text is suboptimal for vision-language learning.

To this end, we propose \name: Bootstrapping Language-Image Pre-training for unified vision-language understanding and generation.
\name~is a new VLP framework which enables a wider range of downstream tasks than existing methods.
It introduces two contributions from the model and data perspective, respectively:

(a) Multimodal mixture of Encoder-Decoder (MED):
a new model architecture for effective multi-task pre-training and flexible transfer learning.
An MED can operate either as a unimodal encoder, or an image-grounded text encoder, or an image-grounded text decoder.
The model is jointly pre-trained with three vision-language objectives: image-text contrastive learning, image-text matching, and image-conditioned language modeling.


(b) Captioning and Filtering (CapFilt):
a new dataset boostrapping method for learning from noisy image-text pairs.
We finetune a pre-trained MED into two modules: a \textit{captioner} to produce synthetic captions given web images,
and a \textit{filter} to remove noisy captions from both the original web texts and the synthetic texts.

\begin{figure}
\vspace{-1ex}
    \centering 
\includegraphics[width=1\columnwidth]{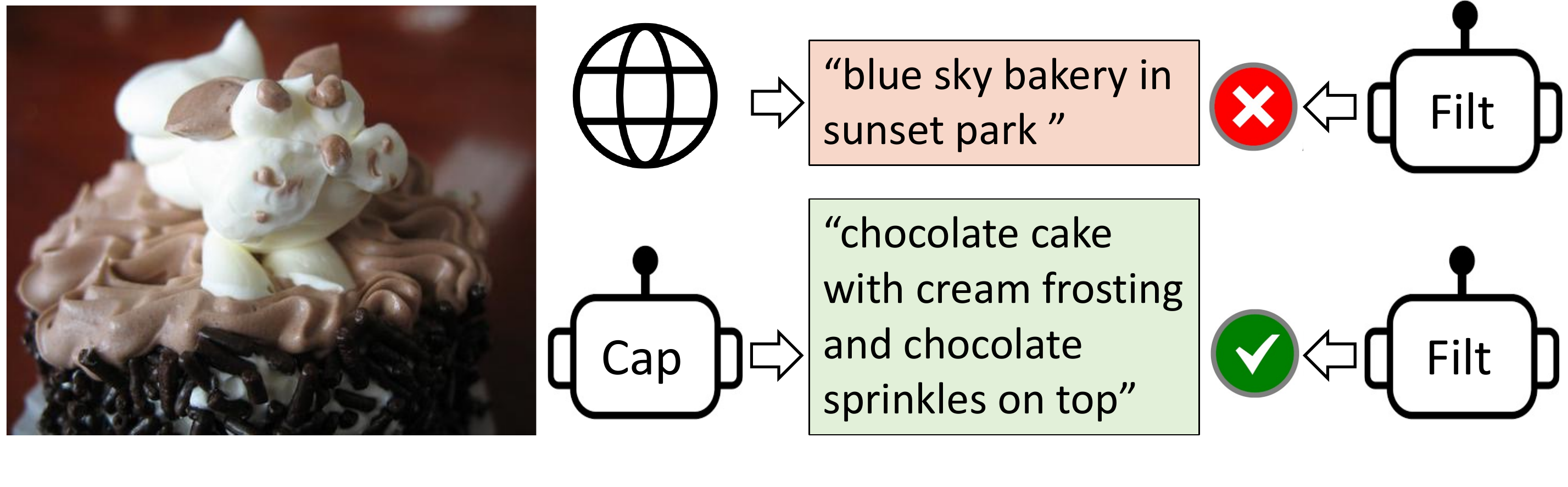}
    \vspace{-5.5ex}
    \caption{We use a Captioner (Cap) to generate synthetic captions for web images, and a Filter (Filt) to remove noisy captions.}
    \label{fig:teaser}
    \vspace{-2.5ex}
\end{figure}

We perform extensive experiments and analysis, and make the following key observations. 
\vspace{-\topsep}
\vspace{-1ex}
\begin{itemize}[leftmargin=*]
	\setlength\itemsep{0pt}
	\item We show that the captioner and the filter work together to achieve substantial performance improvement on various downstream tasks by bootstrapping the captions.  
	We also find that more diverse captions yield larger gains.
	\item \name~achieves state-of-the-art performance on a wide range of vision-language tasks,
	including image-text retrieval, image captioning, visual question answering, visual reasoning, and visual dialog. We also achieve state-of-the-art zero-shot performance when directly transferring our models to two video-language tasks: text-to-video retrieval and videoQA.
\end{itemize}
\vspace{-2.5ex}
\section{Related Work}

\begin{figure*}[!t]
\vspace{-0.5ex}
\centering
  \includegraphics[width=0.98\textwidth]{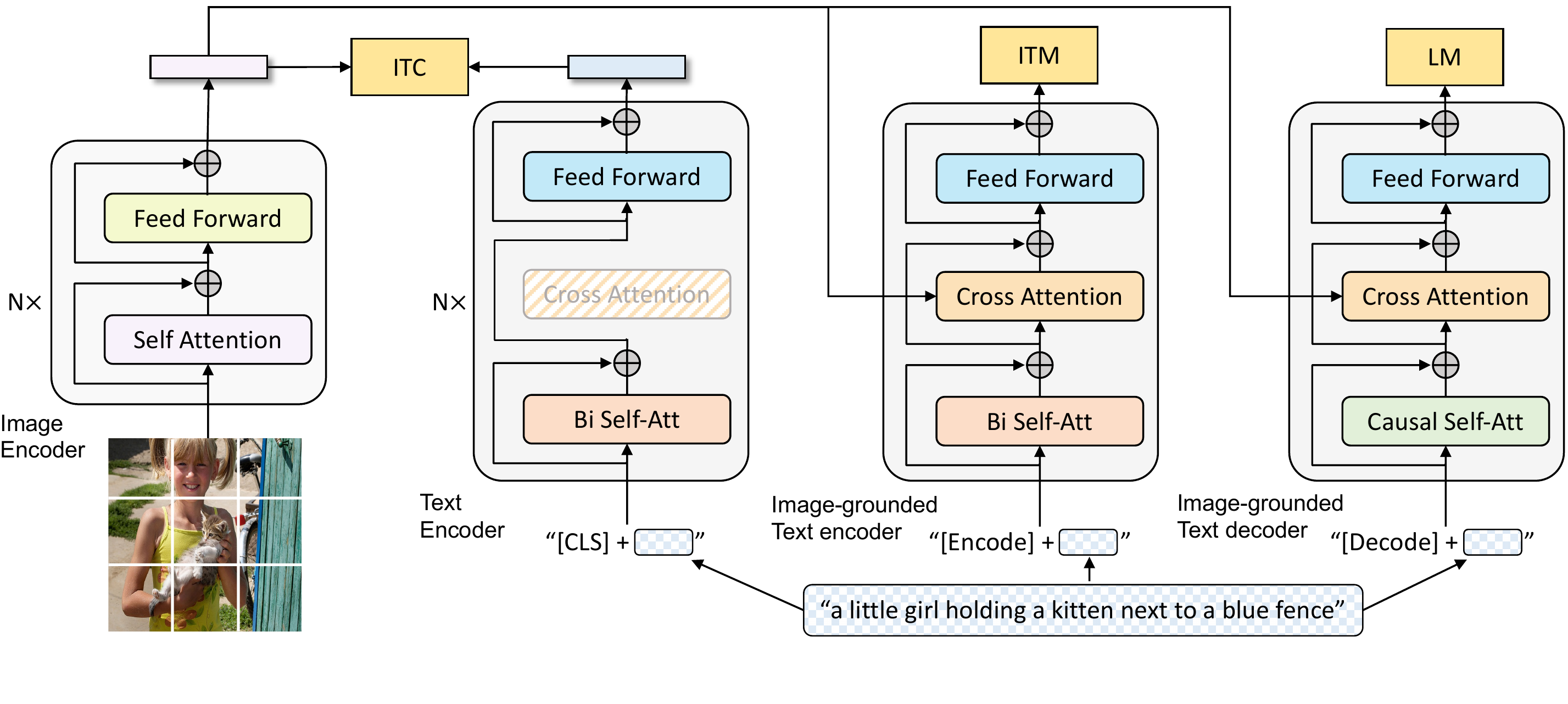}
\vspace{-6ex}
\caption{Pre-training model architecture and objectives of \name~(same parameters have the same color). We propose multimodal mixture of encoder-decoder, a unified vision-language model which can operate in one of the three functionalities:
(1) Unimodal encoder is trained with an image-text contrastive (ITC) loss to align the vision and language representations.
(2) Image-grounded text encoder uses additional cross-attention layers to model vision-language interactions,
and is trained with a image-text matching (ITM) loss to distinguish between positive and negative image-text pairs.
(3) Image-grounded text decoder replaces the bi-directional self-attention layers with causal self-attention layers,
and shares the same cross-attention layers and feed forward networks as the encoder. The decoder is trained with a language modeling (LM) loss to generate captions given images. 
}
\vspace{-2ex}
\label{fig:med}
\end{figure*}

\vspace{-0.5ex}
\subsection{Vision-language Pre-training}
\vspace{-0.5ex}
Vision-language pre-training (VLP) aims to improve performance of downstream vision and language tasks
by pre-training the model on large-scale image-text pairs.
Due to the prohibitive expense of acquiring human-annotated texts,
most methods~\cite{uniter,oscar,ALBEF,simvlm,clip} use image and alt-text pairs crawled from the web~\cite{CC,cc12m,align},
Despite the use of simple rule-based filters,
noise is still prevalent in the web texts.
However, the negative impact of the noise has been largely overlooked, shadowed by the performance gain 
obtained from scaling up the dataset.
Our paper shows that the noisy web texts are suboptimal for vision-language learning,
and proposes CapFilt that utilizes web datasets in a more effective way.

There have been many attempts to unify various vision and language tasks into a single framework~\cite{VLP,VL_T5,simvlm}.
The biggest challenge is to design model architectures that can perform both understanding-based tasks (\eg~image-text retrieval) and generation-based tasks (\eg~image captioning).
Neither encoder-based models~\cite{ALBEF,unimo,clip} nor encoder-decoder models~\cite{VL_T5,simvlm} can excel at both types of tasks,
whereas a single unified encoder-decoder~\cite{VLP} also limits the model's capability.
Our proposed multimodal mixture of encoder-decoder model offers more flexibility and better performance on a wide range of downstream tasks,
in the meantime keeping the pre-training simple and efficient.

\vspace{-1ex}
\subsection{Knowledge Distillation}
\vspace{-0.5ex}
Knowledge distillation (KD)~\cite{kd_hinton} aims to improve the performance of a student model by distilling knowledge from a teacher model.
Self-distillation is a special case of KD where the teacher and student have equal sizes.
It has been shown to be effective for image classification~\cite{noisy_student},
and recently for VLP~\cite{ALBEF}.
Different from mostly existing KD methods which simply enforce the student to have the same class predictions as the teacher,
our proposed CapFilt can be interpreted as a more effective way to perform KD in the context of VLP,
where the captioner distills its knowledge through semantically-rich synthetic captions,
and the filter distills its knowledge by removing noisy captions.

\vspace{-1ex}
\subsection{Data Augmentation}
\vspace{-0.5ex}
While data augmentation (DA) has been widely adopted in computer vision~\cite{DA_survey},
DA for language tasks is less straightforward.
Recently,
generative language models have been used to synthesize examples for various NLP tasks~\cite{DA,DA_AAAI,DA_QA,DA_commonsense}.
Different from these methods which focus on the low-resource language-only tasks,
our method demonstrates the advantage of synthetic captions in large-scale vision-language pre-training.

\vspace{-1.5ex}
\section{Method}
\vspace{-0.5ex}
\label{sec:method}

We propose \name,
a unified VLP framework to learn from noisy image-text pairs.
This section first introduces our new model architecture MED and its pre-training objectives,
and then delineates CapFilt for dataset bootstrapping. 

\vspace{-1ex}
\subsection{Model Architecture}
\label{sec:med}
\vspace{-0.5ex}

We employ a visual transformer~\cite{vit} as our image encoder,
which divides an input image into patches and encodes them as a sequence of embeddings, with an additional \texttt{[CLS]} token to represent the global image feature.
Compared to using pre-trained object detectors for visual feature extraction~\cite{uniter},
using a ViT is more computation-friendly and has been adopted by the more recent methods~\cite{ALBEF,ViLT}.

In order to pre-train a unified model with both understanding and generation capabilities,
we propose multimodal mixture of encoder-decoder (MED),
a multi-task model which can operate in one of the three functionalities:

\vspace{-0.8ex}
(1) \textbf{Unimodal encoder}, which separately encodes image and text. The text encoder is the same as BERT~\cite{bert}, where a \texttt{[CLS]} token is appended to the beginning of the text input to summarize the sentence.

\vspace{-0.8ex}
(2) \textbf{Image-grounded text encoder}, which injects visual information by inserting one additional cross-attention (CA) layer between the self-attention (SA) layer and the feed forward network (FFN) for each transformer block of the text encoder. A task-specific \texttt{[Encode]} token is appended to the text, and the output embedding of \texttt{[Encode]} is used as the multimodal representation of the image-text pair. 

\vspace{-0.8ex}
(3) \textbf{Image-grounded text decoder}, which replaces the bi-directional self-attention layers in the image-grounded text encoder with causal self-attention layers. A \texttt{[Decode]} token is used to signal the beginning of a sequence, and an end-of-sequence token is used to signal its end.

\vspace{-1ex}
\subsection{Pre-training Objectives}
\vspace{-0.5ex}
We jointly optimize three objectives during pre-training,
with two understanding-based objectives and one generation-based objective.
Each image-text pair only requires one forward pass through the computational-heavier visual transformer,
and three forward passes through the text transformer,
where different functionalities are activated to compute the three losses as delineated below.

\noindent\textbf{Image-Text Contrastive Loss} (ITC) activates the unimodal encoder. It aims to align the feature space of the visual transformer and the text transformer by encouraging positive image-text pairs to have similar representations in contrast to the negative pairs. 
It has been shown to be an effective objective for improving vision and language understanding~\cite{clip, ALBEF}.
We follow the ITC loss by~\citet{ALBEF},
where a momentum encoder is introduced to produce features,
and soft labels are created from the momentum encoder as training targets to account for the potential positives in the negative pairs. 

\noindent\textbf{Image-Text Matching Loss} (ITM) activates the image-grounded text encoder.
It aims to learn image-text multimodal representation that captures the fine-grained alignment between vision and language. 
ITM is a binary classification task,
where the model uses an ITM head (a linear layer) to predict whether an image-text pair is positive (matched) or negative (unmatched) given their multimodal feature.
In order to find more informative negatives,
we adopt the hard negative mining strategy by~\citet{ALBEF},
where negatives pairs with higher contrastive similarity in a batch are more likely to be selected to compute the loss.

\noindent\textbf{Language Modeling Loss} (LM) activates the image-grounded text decoder,
which aims to generate textual descriptions given an image.
It optimizes a cross entropy loss which trains the model to maximize the likelihood of the text in an autoregressive manner.
We apply a label smoothing of 0.1 when computing the loss.
Compared to the MLM loss that has been widely-used for VLP,
LM enables the model with the generalization capability to convert visual information into coherent captions.

In order to perform efficient pre-training while leveraging multi-task learning,
the text encoder and text decoder share all parameters except for the SA layers.
The reason is that the differences between the encoding and decoding tasks are best captured by the SA layers. In particular, the encoder employs \textit{bi-directional} self-attention to build representations for the \textit{current} input tokens, while the decoder employs \textit{causal} self-attention to predict \textit{next} tokens.
On the other hand, the embedding layers, CA layers and FFN function similarly between encoding and decoding tasks,
therefore sharing these layers can improve training efficiency while benefiting from multi-task learning,


\begin{figure*}[!ht]
\centering
  \includegraphics[width=\textwidth]{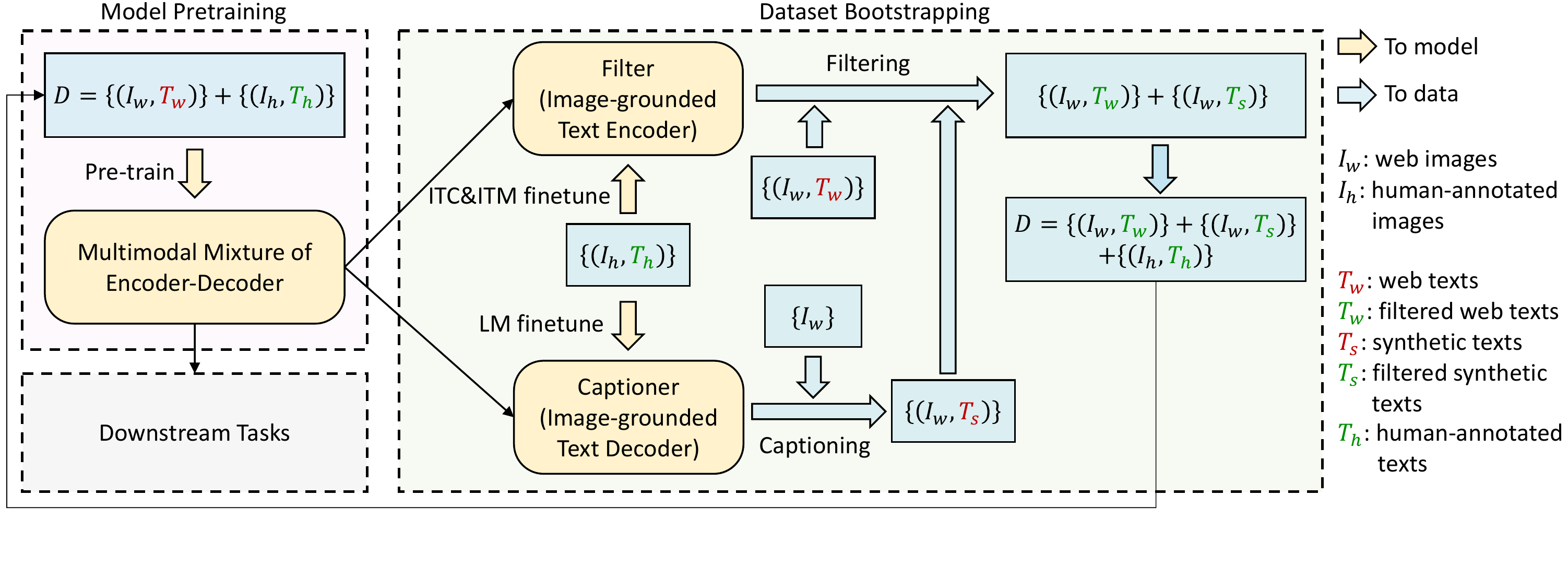}
\vspace{-8ex}
\caption{Learning framework of \name. We introduce a captioner to produce synthetic captions for web images, and a filter to remove noisy image-text pairs.
The captioner and filter are initialized from the same pre-trained model and finetuned individually on a small-scale human-annotated dataset.
The bootstrapped dataset is used to pre-train a new model.
}
\vspace{-2ex}
\label{fig:capfilt}
\end{figure*}

\vspace{-0.5ex}
\subsection{CapFilt}
\vspace{-0.5ex}
Due to the prohibitive annotation cost,
there exist a limited number of high-quality human-annotated image-text pairs $\{(I_h,T_h)\}$ (e.g., COCO~\cite{coco}).
Recent work~\cite{ALBEF,simvlm} utilizes a much larger number of image and alt-text pairs $\{(I_w,T_w)\}$ that are automatically collected from the web.
However, the alt-texts often do not accurately describe the visual content of the images, making them a noisy signal that is suboptimal for learning vision-language alignment. 

We propose Captioning and Filtering (CapFilt),
a new method to improve the quality of the text corpus.
Figure~\ref{fig:capfilt} gives an illustration of CapFilt. 
It introduces two modules: a \textit{captioner} to generate captions given web images,
and a \textit{filter} to remove noisy image-text pairs.
Both the captioner and the filter are initialized from the same pre-trained MED model,
and finetuned individually on the COCO dataset.
The finetuning is a lightweight procedure.

Specifically,
the \emph{captioner} is an image-grounded text decoder.
It is finetuned with the LM objective to decode texts given images. 
Given the web images $I_w$, the captioner generates synthetic captions $T_s$ with one caption per image.
The \emph{filter} is an image-grounded text encoder.
It is finetuned with the ITC and ITM objectives to learn whether a text matches an image.
The filter removes noisy texts in both the original web texts $T_w$ and the synthetic texts $T_s$,
where a text is considered to be noisy if the ITM head predicts it as unmatched to the image.
Finally,
we combine the filtered image-text pairs with the human-annotated pairs to form a new dataset,
which we use to pre-train a new model.

\vspace{-1ex}
\section{Experiments and Discussions}
\label{sec:experiment}
In this section, we first introduce pre-training details. Then we provide a detailed experimental analysis on our method.

\vspace{-1ex}
\subsection{Pre-training Details}
\vspace{-0.5ex}

Our models are implemented in PyTorch~\cite{pytorch} and pre-trained on two 16-GPU nodes.
The image transformer is initialized from ViT pre-trained on ImageNet~\cite{deit,vit},
and the text transformer is initialized from BERT$_\mathrm{base}~$\cite{bert}.
We explore two variants of ViTs: ViT-B/16 and ViT-L/16.
Unless otherwise specified, all results reported in this paper as ``\name'' uses ViT-B.
We pre-train the model for 20 epochs using a batch size of 2880 (ViT-B) / 2400 (ViT-L). 
We use AdamW~\cite{adamw} optimizer with a weight decay of 0.05. 
The learning rate is warmed-up to $3e$-4 (ViT-B) / $2e$-4 (ViT-L) and decayed linearly with a rate of 0.85.
We take random image crops of resolution $224\times224$ during pre-training, and increase the image resolution to $384\times384$ during finetuning.
We use the same pre-training dataset as~\citet{ALBEF} with 14M images in total,
including two human-annotated datasets (COCO and Visual Genome~\cite{VG}),
and three web datasets (Conceptual Captions~\cite{cc12m}, Conceptual 12M~\cite{cc12m}, SBU captions~\cite{sbu}).
We also experimented with an additional web dataset, LAION~\cite{laion}, which contains 115M images with more noisy texts\footnote{We only download images whose shorter edge is larger than 256 pixels from the original LAION400M. Due to the large size of LAION, we only use $1/5$ of it each epoch during pre-training.}.
More details about the datasets can be found in the appendix.

\begin{table*}[!t]
    \small
	\centering	
    \setlength\tabcolsep{4pt}
	\resizebox{\textwidth}{!}{%
	\begin{tabular}	{l | l l | l | C{1cm} c | C{1cm} c | C{1cm} c | C{1cm} c}
	 \toprule	 	
	 \multirow{2}{*}{\makecell[l]{Pre-train \\ dataset}} & \multicolumn{2}{c|}{Bootstrap} & \multirow{2}{*}{\makecell[l]{Vision \\ backbone}} & \multicolumn{2}{c|}{Retrieval-FT (COCO)} & \multicolumn{2}{c|}{Retrieval-ZS (Flickr)} & \multicolumn{2}{c|}{Caption-FT (COCO)} & \multicolumn{2}{c}{Caption-ZS (NoCaps)}\\
     &C &F & & TR@1 & IR@1 & TR@1 & IR@1 & B@4 & CIDEr & CIDEr & SPICE \\
     \midrule
	 \multirow{4}{*}{\makecell[l]{COCO+VG\\+CC+SBU\\(14M imgs)}} & \textcolor{BrickRed}{\xmark} & \textcolor{BrickRed}\xmark  &\multirow{4}{*}{ViT-B/16} &78.4 & 60.7 & 93.9 & 82.1 & 38.0 & 127.8 & 102.2 & 13.9 \\
	  &\textcolor{BrickRed}\xmark & \textcolor{ForestGreen}{\cmark$_B$} & & 79.1 & 61.5 & 94.1 & 82.8 & 38.1 & 128.2 & 102.7 & 14.0 \\
	  & \textcolor{ForestGreen}{\cmark$_B$} & \textcolor{BrickRed}\xmark & & 79.7 & 62.0 & 94.4 & 83.6 & 38.4 & 128.9 & 103.4 & 14.2 \\
     &  \textcolor{ForestGreen}{\cmark$_B$} & \textcolor{ForestGreen}{\cmark$_B$} & & 80.6 & 63.1 & 94.8 & 84.9 & 38.6 & 129.7 & 105.1 & 14.4 \\    

	 \midrule
     \multirow{5}{*}{\makecell[l]{COCO+VG\\+CC+SBU\\+LAION\\(129M imgs)}} & \textcolor{BrickRed}\xmark  & \textcolor{BrickRed}\xmark & \multirow{3}{*}{ViT-B/16} & 79.6 & 62.0 & 94.3  & 83.6 & 38.8 & 130.1 &  105.4 & 14.2 \\
     & \textcolor{ForestGreen}{\cmark$_B$} & \textcolor{ForestGreen}{\cmark$_B$} &  & 81.9  & 64.3 & 96.0 & 85.0 & 39.4 & 131.4 & 106.3 & 14.3 \\ 
     & \textcolor{ForestGreen}{\cmark$_L$} & \textcolor{ForestGreen}{\cmark$_L$} & & 81.2 & 64.1 & 96.0 & 85.5 & 39.7 &  133.3 & 109.6 & 14.7\\ 
     \cmidrule{2-12}
     & \textcolor{BrickRed}\xmark & \textcolor{BrickRed}\xmark & \multirow{2}{*}{ViT-L/16}   & 80.6 &  64.1 & 95.1 & 85.5 & 40.3 & 135.5 &  112.5 & 14.7 \\
     & \textcolor{ForestGreen}{\cmark$_L$} & \textcolor{ForestGreen}{\cmark$_L$} & & 82.4 & 65.1 & 96.7 & 86.7 & 40.4 & 136.7  & 113.2 & 14.8 \\
	\bottomrule
	\end{tabular}
	}
	\vspace{-2ex}
	\caption
	{
	\small	
		Evaluation of the effect of the captioner (C) and filter (F) for dataset bootstrapping.
		Downstream tasks include image-text retrieval and image captioning with finetuning (FT) and zero-shot (ZS) settings.
		TR / IR@1: recall@1 for text retrieval / image retrieval. 
		\textcolor{ForestGreen}{\cmark$_{B/L}$}: captioner or filter uses ViT-B / ViT-L as vision backbone.
	}
	\vspace{-0.5ex}
	\label{tbl:capfilt}
\end{table*}		
\begin{figure*}[!t]
\centering
  \includegraphics[width=\textwidth]{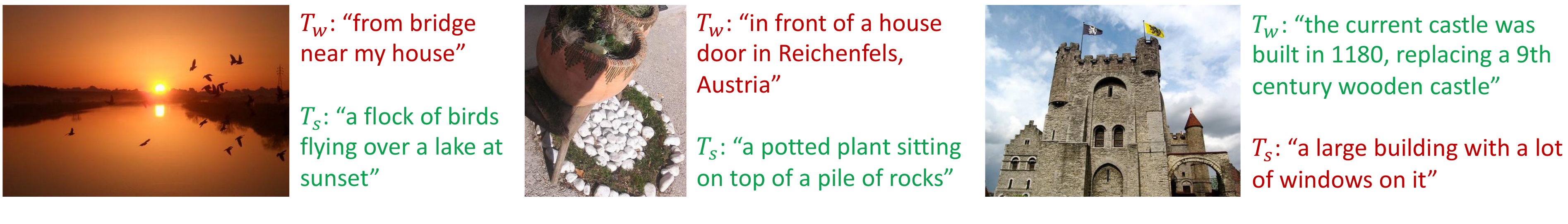}
\vspace{-6ex}
\caption{Examples of the web text $T_w$ and the synthetic text $T_s$. {\color{ForestGreen}Green} texts are accepted by the filter,
whereas {\color{BrickRed}red} texts are rejected.
}
\label{fig:example}
\end{figure*}

\begin{table*}[!t]
	\centering	
	\resizebox{0.95\textwidth}{!}{%
	\begin{tabular}	{l l | C{1cm} c | C{1cm} c | C{1cm} c | C{1cm} c}
	 \toprule	 	
	 \multirow{2}{*}{\makecell[l]{Generation \\ method}} & \multirow{2}{*}{\makecell[l]{Noise \\ ratio}} & \multicolumn{2}{c|}{Retrieval-FT (COCO)} & \multicolumn{2}{c|}{Retrieval-ZS (Flickr)} & \multicolumn{2}{c|}{Caption-FT (COCO)} & \multicolumn{2}{c}{Caption-ZS (NoCaps)}\\
     & & TR@1 & IR@1 & TR@1 & IR@1 & B@4 & CIDEr & CIDEr & SPICE \\
     \midrule
     None    & N.A.    & 78.4 & 60.7 & 93.9 & 82.1 & 38.0 & 127.8 & 102.2 & 13.9 \\
	 Beam    & 19\%  & 79.6 & 61.9 & 94.1 & 83.1 & 38.4 & 128.9 & 103.5 & 14.2 \\  
	 Nucleus & 25\%  & 80.6 & 63.1 & 94.8 & 84.9 & 38.6 & 129.7 & 105.1 & 14.4 \\  

		\bottomrule
	\end{tabular}
	}
	\vspace{-1ex}
	\caption
	{
	\small	
		Comparison between beam search and nucleus sampling for synthetic caption generation. Models are pre-trained on 14M images.
	}
	\label{tbl:generate}
\end{table*}		
\begin{table*}[!t]
	\centering	
   \setlength\tabcolsep{5pt}
	\resizebox{0.95\textwidth}{!}{%
	\begin{tabular}	{l l| C{1cm} c | C{1cm} c | C{1cm} c | C{1cm} c}
	 \toprule	 	
	 \multirow{2}{*}{Layers shared}& \multirow{2}{*}{\#parameters} & \multicolumn{2}{c|}{Retrieval-FT (COCO)} & \multicolumn{2}{c|}{Retrieval-ZS (Flickr)} & \multicolumn{2}{c|}{Caption-FT (COCO)} & \multicolumn{2}{c}{Caption-ZS (NoCaps)}\\
     & & TR@1 & IR@1 & TR@1 & IR@1 & B@4 & CIDEr & CIDEr & SPICE \\
     \midrule
     All & 224M & 77.3 & 59.5 & 93.1 & 81.0 & 37.2 & 125.9 & 100.9 & 13.1 \\
     All except CA & 252M &77.5 & 59.9 & 93.1 & 81.3 & 37.4 & 126.1 & 101.2 & 13.1 \\
	 All except SA & 252M &78.4 & 60.7 & 93.9 & 82.1 & 38.0 & 127.8 & 102.2 & 13.9 \\
	 None & 361M &78.3 & 60.5 & 93.6 & 81.9 & 37.8 & 127.4 & 101.8 & 13.9 \\  
		\bottomrule
	\end{tabular}
	}
	\vspace{-1ex}
	\caption
	{
	\small	
		Comparison between different parameter sharing strategies for the text encoder and decoder during pre-training.
	}
	\vspace{-2ex}
	\label{tbl:med}
\end{table*}		

\vspace{-0.5ex}
\subsection{Effect of CapFilt}
\vspace{-0.5ex}

In Table~\ref{tbl:capfilt},
we compare models pre-trained on different datasets to demonstrate the efficacy of CapFilt on downstream tasks, including image-text retrieval and image captioning with finetuned and zero-shot settings.

When only the captioner or the filter is applied to the dataset with 14M images,
performance improvement can be observed.
When applied together, their effects compliment each other, leading to substantial improvements compared to using the original noisy web texts.

CapFilt can further boost performance with a larger dataset and a larger vision backbone, which verifies its scalability in both the data size and the model size.
Furthermore,
by using a large captioner and filter with ViT-L,
performance of the base model can also be improved.

In Figure~\ref{fig:example},
we show some example captions and their corresponding images,
which qualitatively demonstrate the effect of the captioner to generate new textual descriptions, and the filter to remove noisy captions from both the original web texts and the synthetic texts. 
More examples can be found in the appendix.

\vspace{-0.5ex}
\subsection{Diversity is Key for Synthetic Captions}
\vspace{-0.5ex}
In CapFilt,
we employ nucleus sampling~\cite{nucleus} to generate synthetic captions.
Nucleus sampling is a stochastic decoding method,
where each token is sampled from a set of tokens whose cumulative probability mass exceeds a threshold $p$ ($p=0.9$ in our experiments). 
In Table~\ref{tbl:generate},
we compare it with beam search, a deterministic decoding method which aims to generate captions with the highest probability.
Nucleus sampling leads to evidently better performance,
despite being more noisy as suggested by a higher noise ratio from the filter.
We hypothesis that the reason is that nucleus sampling generates more diverse and surprising captions,
which contain more new information that the model could benefit from.
On the other hand,
beam search tends to generate safe captions that are common in the dataset,
hence offering less extra knowledge.



\vspace{-0.5ex}
\subsection{Parameter Sharing and Decoupling}
\vspace{-0.5ex}

\begin{table*}[!t]
	\centering	
   \setlength\tabcolsep{5pt}
	\resizebox{0.95\textwidth}{!}{%
	\begin{tabular}	{l l | C{1cm} c | C{1cm} c | C{1cm} c | C{1cm} c}
	 \toprule	 	
	 \multirow{2}{*}{\makecell[l]{Captioner \& \\ Filter}} & \multirow{2}{*}{\makecell[l]{Noise \\ ratio}} & \multicolumn{2}{c|}{Retrieval-FT (COCO)} & \multicolumn{2}{c|}{Retrieval-ZS (Flickr)} & \multicolumn{2}{c|}{Caption-FT (COCO)} & \multicolumn{2}{c}{Caption-ZS (NoCaps)}\\
     & & TR@1 & IR@1 & TR@1 & IR@1 & B@4 & CIDEr & CIDEr & SPICE \\
     \midrule
     Share parameters   & 8\%  & 79.8 & 62.2 & 94.3 & 83.7 & 38.4 & 129.0 & 103.5 & 14.2 \\
	 Decoupled & 25\%  & 80.6 & 63.1 & 94.8 & 84.9 & 38.6 & 129.7 & 105.1 & 14.4 \\ \bottomrule
	\end{tabular}
	}
	\vspace{-1ex}
	\caption
	{
	\small	
		Effect of sharing parameters between the captioner and filter.
		Models are pre-trained on 14M images.
	}
	\label{tbl:independent}
\end{table*}		

During pre-training, the text encoder and decoder share all parameters except for the self-attention layers.
In Table~\ref{tbl:med},
we evaluate models pre-trained with different parameter sharing strategies,
where pre-training is performed on the 14M images with web texts.
As the result shows,
sharing all layers except for SA leads to better performance compared to not sharing,
while also reducing the model size thus improveing training efficiency.
If the SA layers are shared, the model's performance would degrade due to the conflict between the encoding task and the decoding task. 

During CapFilt,
the captioner and the filter are end-to-end finetuned individually on COCO.
In Table~\ref{tbl:independent},
we study the effect if the captioner and filter share parameters in the same way as pre-training.
The performance on the downstream tasks decreases,
which we mainly attribute to \textit{confirmation bias}.
Due to parameter sharing,
noisy captions produced by the captioner are less likely to be filtered out by the filter, as indicated by the lower noise ratio (8\% compared to 25\%).

\vspace{-1.5ex}
\section{Comparison with State-of-the-arts}
\label{sec:downstream}
\vspace{-0.5ex}

\begin{table*}[!t]
    \small
	\centering	
 \setlength\tabcolsep{4pt}
	\resizebox{\textwidth}{!}{%
	\begin{tabular}	{l  l |  c  c  c  c  c  c | c  c  c  c  c  c }
		\toprule	 	
	 \multirow{2}{*}{Method} & Pre-train & \multicolumn{6}{c|}{COCO (5K test set)} & \multicolumn{6}{c}{Flickr30K (1K test set)} \\
	 & \# Images &  \multicolumn{3}{c}{TR}& \multicolumn{3}{c|}{IR} &  \multicolumn{3}{c}{TR}& \multicolumn{3}{c}{IR}\\
	 \midrule
	& & R@1 &R@5&R@10& R@1 &R@5&R@10& R@1 &R@5&R@10& R@1 &R@5&R@10\\
	UNITER~\cite{uniter} & 4M &65.7 &88.6 &93.8 &52.9 &79.9 &88.0 & 87.3 & 98.0 &99.2 &75.6 &94.1 &96.8 \\
	VILLA~\cite{villa} & 4M & - & - & - & - & - & -  &87.9 & 97.5 & 98.8 & 76.3 & 94.2 & 96.8 \\
	OSCAR~\cite{oscar} & 4M & 70.0 & 91.1 & 95.5 &  54.0 & 80.8 & 88.5 & - & - & - & - & - & - \\
	UNIMO~\cite{unimo} & 5.7M &  - & - & - & - & - & - &  89.4 & 98.9 & 99.8 & 78.0 & 94.2 & 97.1 \\
	ALIGN~\cite{align} & 1.8B & 77.0 & 93.5 & 96.9 & 59.9 & 83.3 & 89.8 & 95.3 &  99.8 & 100.0 & 84.9 & 97.4 & 98.6 \\
	ALBEF~\cite{ALBEF} & 14M &
	77.6 & 94.3& 97.2 & 60.7& 84.3 & 90.5 & 95.9 & 99.8 & 100.0 & 85.6& 97.5 & 98.9\\
    \midrule
    \name & 14M & 80.6 & 95.2 & 97.6  & 63.1 & 85.3 & 91.1 & 96.6 & 99.8 & \textbf{100.0} & 87.2 & 97.5 & 98.8 \\
    \name & 129M & \textbf{81.9} & 95.4 & 97.8  & \textbf{64.3} & 85.7 & 91.5 & \textbf{97.3} & \textbf{99.9} & \textbf{100.0} & 87.3 & 97.6 & \textbf{98.9} \\  
    \name$_\text{CapFilt-L}$ & 129M & 81.2 & \textbf{95.7} & \textbf{97.9}  & 64.1 & \textbf{85.8} & \textbf{91.6} & 97.2 & \textbf{99.9} & \textbf{100.0} & \textbf{87.5} & \textbf{97.7} & \textbf{98.9} \\
    \midrule
    \name$_\text{ViT-L}$ & 129M & 82.4 & 95.4 & 97.9  & 65.1 & 86.3 & 91.8 & 97.4 & 99.8 & 99.9 & 87.6 & 97.7 & 99.0 \\    
		\bottomrule
	\end{tabular}
	}
	\vspace{-2ex}
	\caption
	{
	\small	
		Comparison with state-of-the-art image-text retrieval methods,
		finetuned on COCO and Flickr30K datasets.
		\name$_\text{CapFilt-L}$ pre-trains a model with ViT-B backbone using a dataset bootstrapped by captioner and filter with ViT-L.
	}
	\vspace{-2ex}
	\label{tbl:retrieval}
\end{table*}		
\begin{table}[!t]
    \small
	\centering	
 \setlength\tabcolsep{3pt}
	\resizebox{1.0\columnwidth}{!}{%
	\begin{tabular}	{l  l |  c  c  c  c  c  c  }
		\toprule	 	
	 \multirow{2}{*}{Method} & Pre-train & \multicolumn{6}{c}{Flickr30K (1K test set)}\\
	 & \# Images &  \multicolumn{3}{c}{TR}& \multicolumn{3}{c}{IR} \\
	 \midrule
	& & R@1 &R@5&R@10& R@1 &R@5&R@10 \\
	CLIP & 400M & 88.0 & 98.7 & 99.4 & 68.7 & 90.6 & 95.2 \\
	ALIGN & 1.8B& 88.6 &98.7& 99.7& 75.7& 93.8& 96.8 \\
	
	ALBEF & 14M & 94.1 & 99.5& 99.7 &82.8 & 96.3 & 98.1 \\
	\midrule
	\name & 14M & 94.8 & 99.7 & \textbf{100.0} & 84.9 & 96.7 & 98.3 \\
    \name & 129M & \textbf{96.0} & \textbf{99.9} & \textbf{100.0} & 85.0 & \textbf{96.8} & 98.6\\	
    \name$_\text{CapFilt-L}$ & 129M & \textbf{96.0} & \textbf{99.9} & \textbf{100.0} & \textbf{85.5} & \textbf{96.8} & \textbf{98.7}\\	
    \midrule 
    \name$_\text{ViT-L}$& 129M & 96.7 & 100.0 & 100.0 & 86.7 & 97.3 & 98.7\\   
	\bottomrule
	\end{tabular}
 	}
 	 \vspace{-2ex}
	\caption
	{
	\small	
		Zero-shot image-text retrieval results on Flickr30K.
	}
	\label{tbl:retrieval_zs}
	\vspace{-4.5ex}
\end{table}		

In this section,
we compare \name~to existing VLP methods on a wide range of vision-language downstream tasks\footnote{we omit SNLI-VE from the benchmark because its test data has been reported to be noisy~\cite{ve-2.0}}.
Next we briefly introduce each task and finetuning strategy.
More details can be found in the appendix.

\vspace{-1ex}
\subsection{Image-Text Retrieval}
\vspace{-0.5ex}
We evaluate \name~for both image-to-text retrieval (TR) and text-to-image retrieval (IR) on COCO and Flickr30K~\cite{flickr} datasets.
We finetune the pre-trained model using ITC and ITM losses.
To enable faster inference speed, we follow~\citet{ALBEF} and first select $k$ candidates based on the image-text feature similarity,
and then rerank the selected candidates based on their pairwise ITM scores.
We set $k=256$ for COCO and $k=128$ for Flickr30K.

As shown in Table~\ref{tbl:retrieval},
\name~achieves substantial performance improvement compared with existing methods.
Using the same 14M pre-training images,
\name~outperforms the previous best model ALBEF by +2.7\% in average recall@1 on COCO.
We also perform zero-shot retrieval by directly transferring the model finetuned on COCO to Flickr30K.
The result is shown in Table~\ref{tbl:retrieval_zs},
where \name~also outperforms existing methods by a large margin.

\vspace{-1ex}
\subsection{Image Captioning}
\vspace{-0.5ex}
\begin{table*}[!t]
	\centering	
    \setlength\tabcolsep{5pt}
	\resizebox{1\textwidth}{!}{%
	\begin{tabular}	{l  l |  c  c  c  c  c  c  c  c | c  c }
		\toprule	 	
	 \multirow{3}{*}{Method}& \multirow{3}{*}{\makecell[l]{Pre-train \\ \#Images}} &  \multicolumn{8}{c|}{NoCaps validation} & 
	 \multicolumn{2}{c}{COCO Caption}\\
	 & & \multicolumn{2}{c}{in-domain} & \multicolumn{2}{c}{near-domain} & \multicolumn{2}{c}{out-domain} & \multicolumn{2}{c|}{overall} & \multicolumn{2}{c}{Karpathy test}\\
	 & & C & S & C & S & C & S & C & S & B@4 & C\\
	  \midrule
	   Enc-Dec~\cite{cc12m} & 15M & 92.6 & 12.5 & 88.3 & 12.1 & 94.5 & 11.9 & 90.2 & 12.1 & - &110.9\\
	   VinVL\dag~\cite{vinvl} & 5.7M & 103.1 & 14.2 & 96.1 & 13.8 & 88.3 & 12.1 & 95.5 & 13.5 &  38.2 & 129.3 \\
	   LEMON$_\mathrm{base}$\dag~\cite{lemon} & 12M & 104.5 & 14.6 & 100.7 & 14.0 & 96.7 & 12.4 & 100.4 & 13.8 & - & - \\
	   LEMON$_\mathrm{base}$\dag~\cite{lemon} & 200M & 107.7 & 14.7 & 106.2 & 14.3 & 107.9 & 13.1 & 106.8 & 14.1 & \textbf{40.3} & \textbf{133.3} \\

	   \midrule
	   \name &14M & 111.3 & 15.1 & 104.5 & 14.4 & 102.4 & 13.7  & 105.1 & 14.4  &38.6 & 129.7 \\
	   \name & 129M & 109.1 & 14.8 & 105.8 & 14.4 & 105.7 & 13.7 & 106.3 & 14.3 & 39.4 & 131.4 \\
	   \name$_\text{CapFilt-L}$ & 129M & \textbf{111.8} & \textbf{14.9} & \textbf{108.6} & \textbf{14.8} & \textbf{111.5} & \textbf{14.2} & \textbf{109.6} & \textbf{14.7} & 39.7 & \textbf{133.3} \\	   
	   
	   \midrule
	   \midrule
	   LEMON$_\mathrm{large}$\dag~\cite{lemon} & 200M & 116.9 & 15.8 & 113.3 & 15.1 & 111.3 & 14.0 & 113.4  &15.0 &40.6 & 135.7 \\	   
	   \textcolor{lightgray}{SimVLM$_\mathrm{huge}$}~\cite{simvlm} & \textcolor{lightgray}{1.8B} & \textcolor{lightgray}{113.7} & \textcolor{lightgray}{-} & \textcolor{lightgray}{110.9} & \textcolor{lightgray}{-} & \textcolor{lightgray}{115.2} & \textcolor{lightgray}{-} & \textcolor{lightgray}{112.2} & \textcolor{lightgray}{-} & \textcolor{lightgray}{40.6} & \textcolor{lightgray}{143.3}\\	   
	   \name$_\text{ViT-L}$ & 129M & 114.9 & 15.2 & 112.1 & 14.9 & 115.3 & 14.4 & 113.2 & 14.8 & 40.4 & 136.7 \\	   
		\bottomrule

	\end{tabular}
 	}
 \vspace{-2ex}
	\caption
	{
	\small	
		Comparison with state-of-the-art image captioning methods on NoCaps and COCO Caption.
		All methods optimize the cross-entropy loss during finetuning. C: CIDEr, S: SPICE, B@4: BLEU@4.	
		\name$_\text{CapFilt-L}$ is pre-trained on a dataset bootstrapped by captioner and filter with ViT-L.
		VinVL\dag~and LEMON\dag~require an object detector pre-trained on 2.5M images with human-annotated bounding boxes and high resolution (800$\times$1333) input images. 
		SimVLM$_\mathrm{huge}$ uses 13$\times$ more training data and a larger vision backbone than ViT-L.
	}
	\vspace{-1ex}
	\label{tbl:caption}
\end{table*}		
We consider two datasets for image captioning: NoCaps~\cite{nocaps} and COCO,
both evaluated using the model finetuned on COCO with the LM loss.
Similar as~\citet{simvlm},
we add a prompt ``a picture of'' at the beginning of each caption,
which leads to slightly better results.
As shown in Table~\ref{tbl:caption},
\name~with 14M pre-training images substantially outperforms methods using a similar amount of pre-training data.
\name~with 129M images achieves competitive performance as LEMON with 200M images.
Note that LEMON requires a computational-heavy pre-trained object detector and higher resolution (800$\times$1333) input images,
leading to substantially slower inference time than the detector-free \name~which uses lower resolution (384$\times$384) input images.

\vspace{-1ex}
\subsection{Visual Question Answering (VQA)}
\vspace{-0.5ex}
\begin{figure}[!t]
\centering
  \includegraphics[width=0.98\columnwidth]{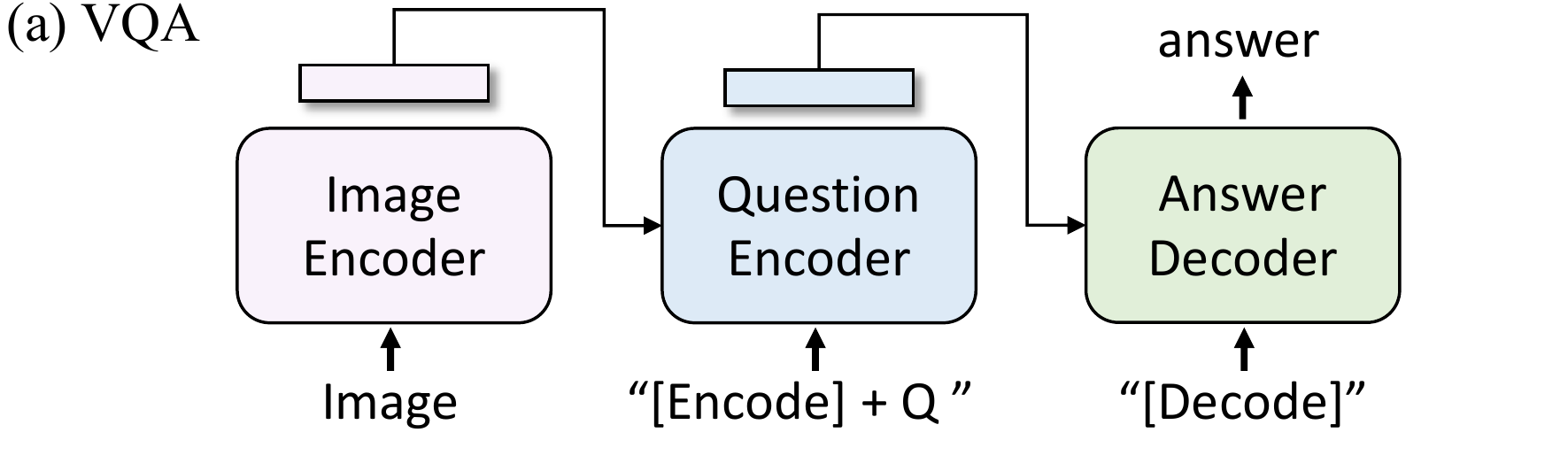}
 \includegraphics[width=0.98\columnwidth]{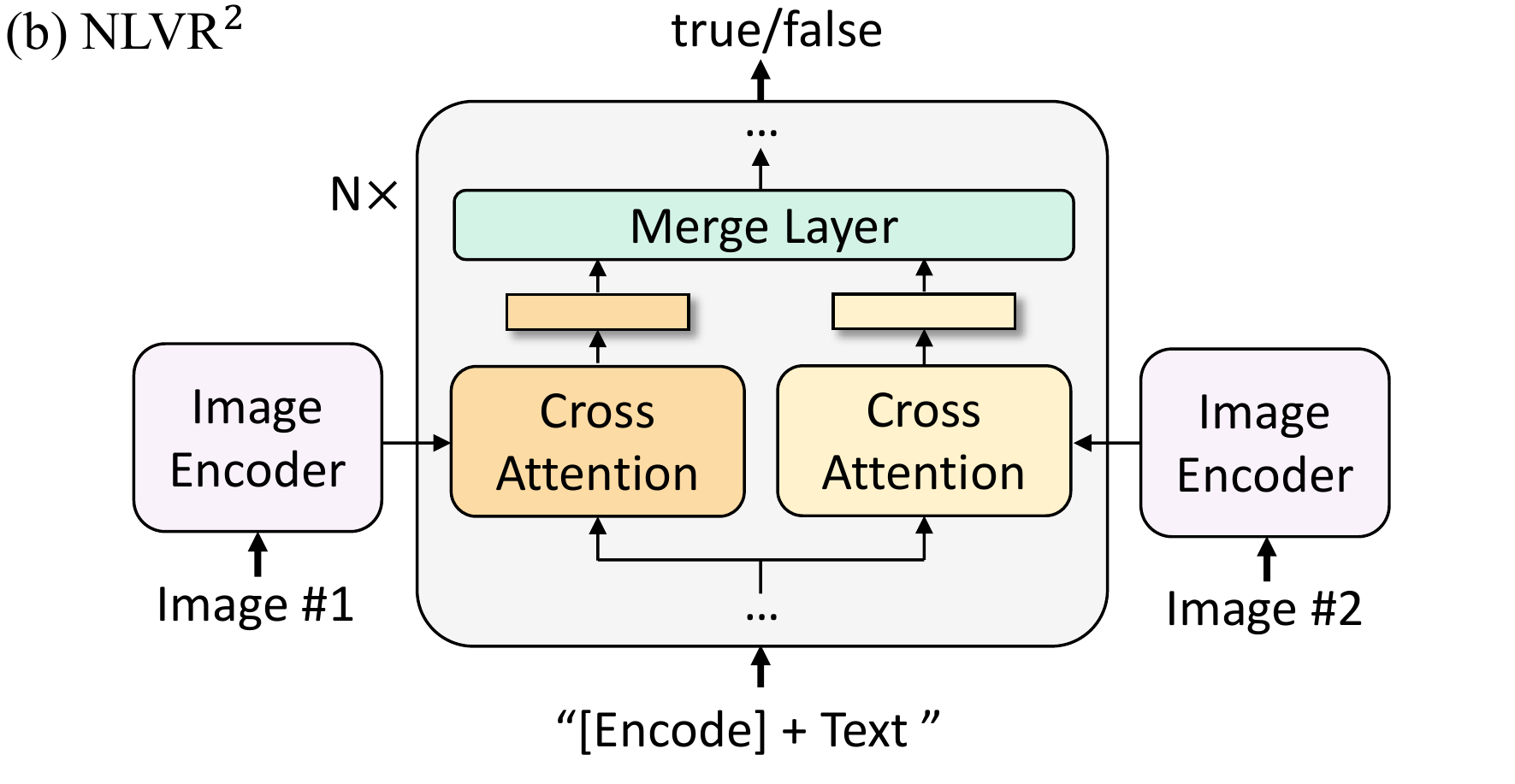}
  \includegraphics[width=0.98\columnwidth]{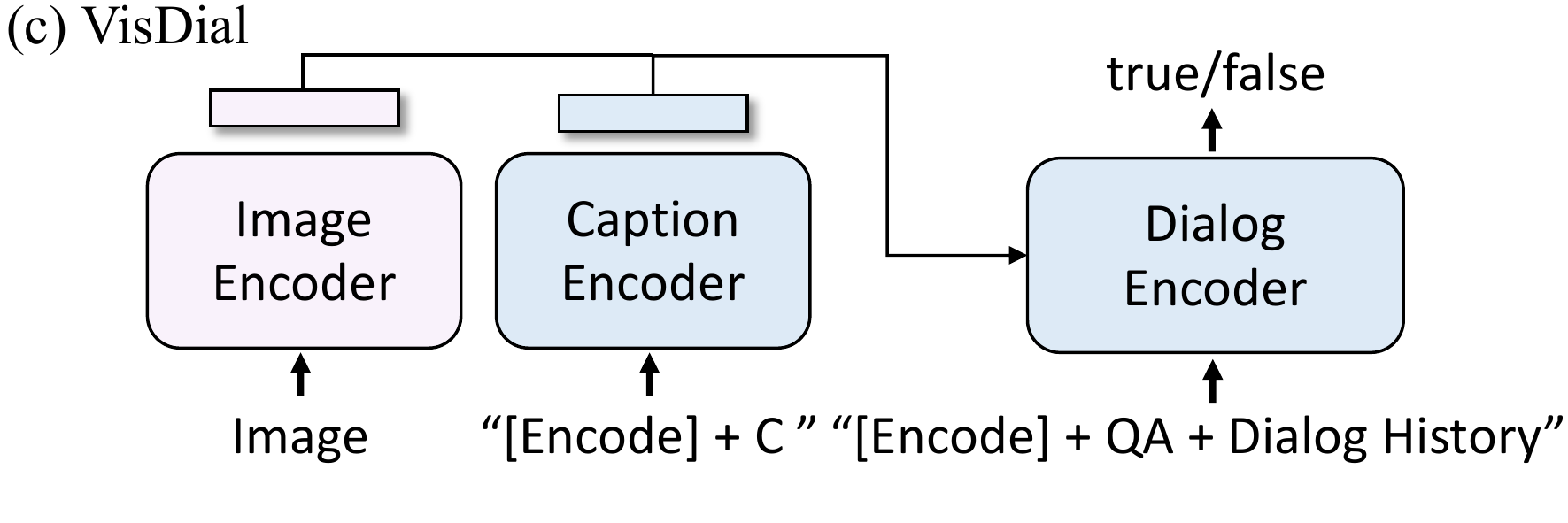}
  \vspace{-3ex}
\caption{Model architecture for the downstream tasks. Q: question; C: caption; QA: question-answer pair.
}
\vspace{-3ex}
\label{fig:vqa_visdial_nlvr}
\end{figure}

\begin{table}[!t]
	\centering	
    \setlength\tabcolsep{5pt}
	\resizebox{1\columnwidth}{!}{%
	\begin{tabular}	{l l  |  c  c  c  c}
		\toprule	 	
	 \multirow{2}{*}{Method}& \multirow{2}{*}{\makecell[l]{Pre-train \\ \#Images}}&  \multicolumn{2}{c}{VQA} & \multicolumn{2}{c}{NLVR$^2$} \\
	  & & test-dev & test-std & dev & test-P \\
	  \midrule
	  LXMERT& 180K & 72.42 & 72.54 & 74.90 & 74.50  \\
	  UNITER& 4M & 72.70 & 72.91 & 77.18 & 77.85 \\
	   VL-T5/BART & 180K & - & 71.3 & - & 73.6 \\
	   OSCAR& 4M & 73.16 & 73.44 & 78.07 & 78.36 \\
	   SOHO& 219K & 73.25 &73.47 &76.37 &77.32\\
	   VILLA& 4M & 73.59 & 73.67 & 78.39 & 79.30\\
	   UNIMO & 5.6M & 75.06 & 75.27 & - & - \\
	   ALBEF & 14M & 75.84 & 76.04 & \textcolor{lightgray}{82.55} & \textcolor{lightgray}{83.14}\\ 
	   SimVLM$_\mathrm{base}$\dag & 1.8B & 77.87 & 78.14 & 81.72 & 81.77 \\
	   \midrule
	   \name &14M & 77.54 & 77.62 & \textbf{82.67} & 82.30 \\
	   \name & 129M & 78.24 & 78.17 & 82.48 & \textbf{83.08} \\
	   \name$_\text{CapFilt-L}$ & 129M & \textbf{78.25} & \textbf{78.32} & 82.15 & 82.24 \\
	\bottomrule
	\end{tabular}
 	}
 \vspace{-2ex}
	\caption
	{
	\small	
		Comparison with state-of-the-art methods on VQA and NLVR$^2$. 
		ALBEF performs an extra pre-training step for NLVR$^2$.
		SimVLM\dag~uses 13$\times$ more training data and a larger vision backbone (ResNet+ViT) than \name.
	}
	\label{tbl:vqa_nlvr}
	\vspace{-4ex}
\end{table}

VQA~\cite{VQA} requires the model to predict an answer given an image and a question.
Instead of formulating VQA as a multi-answer classification task~\cite{uniter,oscar},
we follow~\citet{ALBEF} and consider it as an answer generation task,
which enables open-ended VQA.
As shown in Figure~\ref{fig:vqa_visdial_nlvr}(a),
during finetuning,
we rearrange the pre-trained model,
where an image-question is first encoded into multimodal embeddings and then given to an answer decoder.
The VQA model is finetuned with the LM loss using ground-truth answers as targets.

The results are shown in Table~\ref{tbl:vqa_nlvr}.
Using 14M images, \name~outperforms ALBEF by +1.64\% on the test set.
Using 129M images, \name~achieves better performance than SimVLM which uses $13\times$ more pre-training data and a larger vision backbone with an additional convolution stage.

\vspace{-1ex}
\subsection{Natural Language Visual Reasoning (NLVR$^2$)}
\vspace{-0.5ex}
NLVR$^2$~\cite{NLVR} asks the model to predict whether a sentence describes a pair of images.
In order to enable reasoning over two images,
we make a simple modification to our pre-trained model which leads to a more computational-efficient architecture than previous approaches~\cite{ALBEF,simvlm}.
As shown in Figure~\ref{fig:vqa_visdial_nlvr}(b),
for each transformer block in the image-grounded text encoder,
there exist two cross-attention layers to process the two input images,
and their outputs are merged and fed to the FFN.
The two CA layers are intialized from the same pre-trained weights.
The merge layer performs simple average pooling in the first 6 layers of the encoder,
and performs concatenation followed by a linear projection in layer 6-12.
An MLP classifier is applied on the output embedding of the \texttt{[Encode]} token.
As shown in Table~\ref{tbl:vqa_nlvr},
\name~outperforms all existing methods except for ALBEF which performs an extra step of customized pre-training.
Interestingly,
performance on NLVR$^2$ does not benefit much from additional web images,
possibly due to the domain gap between web data and downstream data.

\vspace{-1ex}
\subsection{Visual Dialog (VisDial)}
\vspace{-0.5ex}
VisDial~\cite{VisDial} extends VQA in a natural conversational setting,
where the model needs to predict an answer not only based on the image-question pair, but also considering the dialog history and the image's caption.
We follow the discriminative setting where the model ranks a pool of answer candidates~\cite{ReDAN,VD-BERT,visdial_eccv}.
As shown in Figure~\ref{fig:vqa_visdial_nlvr}(c),
we concatenate image and caption embeddings,
and pass them to the dialog encoder through cross-attention.
The dialog encoder is trained with the ITM loss to discriminate whether the answer is true or false for a question,
given the entire dialog history and the image-caption embeddings.
As shown in Table~\ref{tbl:visdial},
our method achieves state-of-the-art performance on VisDial v1.0 validation set.

\vspace{-1ex}
\subsection{Zero-shot Transfer to Video-Language Tasks}
\vspace{-0.5ex}

\begin{table}[!t]
    \small
	\centering	
    \setlength\tabcolsep{5pt}
	\resizebox{1\columnwidth}{!}{%
	\begin{tabular}	{l   |  c  c  c  c c }
		\toprule	 	
	 Method & MRR$\uparrow$ & R@1$\uparrow$ & R@5$\uparrow$ & R@10$\uparrow$ & MR$\downarrow$  \\
		\midrule
        VD-BERT & 67.44 & 54.02 & 83.96 & 92.33 & 3.53 \\
        VD-ViLBERT\dag & 69.10 & 55.88 & 85.50 & 93.29 & 3.25\\
        \name & \textbf{69.41} & \textbf{56.44} & \textbf{85.90} & \textbf{93.30}& \textbf{3.20}\\
		\bottomrule
	\end{tabular}
 	}
 \vspace{-2ex}
	\caption
	{
	\small	
		Comparison with state-of-the-art methods on VisDial v1.0 validation set. VD-ViLBERT\dag~\cite{visdial_eccv} pre-trains ViLBERT~\cite{ViLBERT} with additional VQA data.
	}
	\vspace{-3ex}
	\label{tbl:visdial}
\end{table}		
\begin{table}[!t]

 \resizebox{1\columnwidth}{!}{
	\centering	
      \setlength\tabcolsep{4pt}
		\begin{tabular}	{l  |  c c c c}
		\toprule
		Method & R1$\uparrow$ & R5$\uparrow$ & R10$\uparrow$ & MdR$\downarrow$\\
		\midrule
		\multicolumn{3}{l}{\textit{zero-shot}}\\
		\midrule
		ActBERT~\cite{zhu2020actbert} & 8.6 & 23.4 & 33.1 & 36\\
		SupportSet~\cite{patrick2021supportset} & 8.7 & 23.0 & 31.1 & 31\\
        MIL-NCE~\cite{miech2020end} & 9.9 & 24.0 & 32.4 & 29.5\\ VideoCLIP~\cite{xu2021videoclip} & 10.4 & 22.2 & 30.0 & -\\
        FiT~\cite{FiT} & 18.7 & 39.5 & 51.6 & 10 \\ 
        \name & \textbf{43.3} & \textbf{65.6} & \textbf{74.7} & \textbf{2} \\
        \midrule
        \multicolumn{3}{l}{\textit{finetuning}}\\
        \midrule
        ClipBERT~\cite{lei2021less} & 22.0 & 46.8 & 59.9 & 6 \\
        VideoCLIP~\cite{xu2021videoclip} & 30.9 & 55.4  & 66.8 & - \\
        \bottomrule
		\end{tabular}
 		}
 	\vspace{-2ex}
    \caption
	{Comparisons with state-of-the-art methods for text-to-\textbf{video} retrieval on the 1k test split of the MSRVTT dataset.}
	\vspace{-1ex}
	\label{tbl:video_retrieval}
\end{table}		
\begin{table}[!t]

 \resizebox{1\columnwidth}{!}{
	\centering	
     \setlength\tabcolsep{5pt}
		\begin{tabular}	{l  |  c c }
		\toprule
		Method & MSRVTT-QA& MSVD-QA\\
		\midrule
		\multicolumn{3}{l}{\textit{zero-shot}}\\
		\midrule		
		VQA-T~\cite{justask} &  2.9 & 7.5 \\
		\name & 19.2 & 35.2 \\ 
		\midrule
		\multicolumn{3}{l}{\textit{finetuning}}\\
		\midrule
		HME~\cite{fan2019heterogeneous}  & 33.0 & 33.7 \\
    	HCRN~\cite{le2020hierarchical}  & 35.6 & 36.1 \\
		VQA-T~\cite{justask} & 41.5 & 46.3\\		
        \bottomrule
		\end{tabular}
		}
    \vspace{-2ex}		
    \caption
	{Comparisons with state-of-the-art methods for  \textbf{video} question answering. We report top-1 test accuracy on two datasets.}
	\vspace{-3ex}
	\label{tbl:video_qa}
\end{table}

Our image-language model has strong generalization ability to video-language tasks.
In Table~\ref{tbl:video_retrieval} and Table~\ref{tbl:video_qa},
we perform zero-shot transfer to \textit{text-to-video retrieval} and \textit{video question answering},
where we directly evaluate the models trained on COCO-retrieval and VQA, respectively.
To process video input,
we uniformly sample $n$ frames per video ($n=8$ for retrieval and $n=16$ for QA),
and concatenate the frame features into a single sequence.
Note that this simple approach ignores all temporal information.

Despite the domain difference and lack of temporal modeling,
our models achieve state-of-the-art performance on both video-language tasks.
For text-to-video retrieval,
zero-shot \name~even outperforms models finetuned on the target video dataset by +12.4\% in recall@1.
Further performance improvement can be achieved if the \name~model is used to initialize a video-language model with temporal modeling (\eg~replace our ViT with a TimeSformer~\cite{timesformer}) and finetuned on video data.

\begin{table*}[!t]
    \small
	\centering	
	\begin{tabular}	{l l | C{1cm} c | C{1cm} c | C{1cm} c | C{1cm} c}
	 \toprule	 	
	 \multirow{2}{*}{\makecell[l]{CapFilt}} & \multirow{2}{*}{\#Texts} & \multicolumn{2}{c|}{Retrieval-FT (COCO)} & \multicolumn{2}{c|}{Retrieval-ZS (Flickr)} & \multicolumn{2}{c|}{Caption-FT (COCO)} & \multicolumn{2}{c}{Caption-ZS (NoCaps)}\\
     & & TR@1 & IR@1 & TR@1 & IR@1 & B@4 & CIDEr & CIDEr & SPICE \\
     \midrule
     No    &  15.3M & 78.4 & 60.7 & 93.9 & 82.1 & 38.0 & 127.8 & 102.2 & 13.9 \\
	 No    &  24.7M  & 78.3 & 60.5 & 93.7 & 82.2 & 37.9 & 127.7 & 102.1 & 14.0 \\  
	 Yes   &  24.7M & 80.6 & 63.1 & 94.8 & 84.9 & 38.6 & 129.7 & 105.1 & 14.4 \\  

		\bottomrule
	\end{tabular}
	\vspace{-1ex}
	\caption
	{
	\small	
		The original web texts are replicated to have the same number of samples per epoch as the bootstrapped dataset. Results verify that the improvement from CapFilt is not due to longer training time.
	}
	\label{tbl:longer}
\end{table*}		

\begin{table*}[!t]
    \small
	\centering	
	\begin{tabular}	{l | C{1cm} c | C{1cm} c | C{1cm} c | C{1cm} c}
	 \toprule	 	
	 \multirow{2}{*}{\makecell[l]{Continue}} &  \multicolumn{2}{c|}{Retrieval-FT (COCO)} & \multicolumn{2}{c|}{Retrieval-ZS (Flickr)} & \multicolumn{2}{c|}{Caption-FT (COCO)} & \multicolumn{2}{c}{Caption-ZS (NoCaps)}\\
     & TR@1 & IR@1 & TR@1 & IR@1 & B@4 & CIDEr & CIDEr & SPICE \\
     \midrule
	 Yes  & 80.6 & 63.0 & 94.5 & 84.6 & 38.5 & 129.9 & 104.5 & 14.2\\  
	 No  & 80.6 & 63.1 & 94.8 & 84.9 & 38.6 & 129.7 & 105.1 & 14.4 \\  

		\bottomrule
	\end{tabular}
	\vspace{-1ex}
	\caption
	{
	\small	
		Continue training the pre-trained model offers less gain compared to training a new model with the bootstrapped dataset.
	}
\vspace{-2ex}
	\label{tbl:continue}
\end{table*}		

\vspace{-0.5ex}
\section{Additional Ablation Study}

In this section,
we provide additional ablation experiments on CapFilt.

\noindent\textbf{Improvement with CapFilt is not due to longer training.}
Since the bootstrapped dataset contains more texts than the original dataset,
training for the same number of epochs takes longer with the bootstrapped dataset.
To verify that the effectiveness of CapFilt is not due to longer training,
we replicate the web text in the original dataset so that it has the same number of training samples per epoch as the bootstrapped dataset.
As shown in Table~\ref{tbl:longer}, 
longer training using the noisy web texts does not improve performance.

\noindent\textbf{A new model should be trained on the bootstrapped dataset.}
The bootstrapped dataset is used to pre-train a new model.
We investigate the effect of continue training from the previous pre-trained model, using the bootstrapped dataset. 
Table~\ref{tbl:continue} hows that continue training does not help.
This observation agrees with the common practice in knowledge distillation,
where the student model cannot be initialized from the teacher.

\vspace{-1ex}
\section{Conclusion}
\vspace{-0.5ex}
We propose \name,
a new VLP framework with state-of-the-art performance on a wide range of downstream vision-language tasks,
including understanding-based and generation-based tasks.
\name~pre-trains a multimodal mixture of encoder-decoder model using a dataset bootstrapped from large-scale noisy image-text pairs by injecting diverse synthetic captions and removing noisy captions.
Our bootstrapped dataset are released to facilitate future vision-language research.

There are a few potential directions that can further enhance the performance of \name:
(1) Multiple rounds of dataset bootstrapping;
(2) Generate multiple synthetic captions per image to further enlarge the pre-training corpus;
(3) Model ensemble by training multiple different captioners and filters and combining their forces in CapFilt.
We hope that our paper motivates future work to focus on making improvements in both the model aspect and the data aspect,
the bread and butter of vision-language research.

\bibliography{bib}
\bibliographystyle{icml2022}

\clearpage

\appendix

\section{Downstream Task Details}

Table~\ref{tbl:paramter} shows the hyperparameters that we use for finetuning on the downstream vision-language tasks.
All tasks uses AdamW optimizer with a weight decay of 0.05 and a cosine learning rate schedule.
We use an image resolution of $384\times384$,
except for VQA where we follow~\citet{simvlm} and use $480\times480$ images.
Next we delineate the dataset details.

\noindent\textbf{Image-Text Retrieval.}
We use the Karpathy split~\cite{karpathy} for both COCO and Flickr30K.
COCO contains 113/5k/5k images for train/validation/test,
and Flickr30K contains 29k/1k/1k images for train/validation/test.

\noindent\textbf{Image Captioning.}
We finetune on COCO's Karpathy train split,
and evaluate on COCO's Karpathy test split and NoCaps validation split.
During inference,
we use beam search with a beam size of 3,
and set the maximum generation length as 20.

\noindent\textbf{VQA.}
We experiment with the VQA2.0 dataset~\cite{VQA2}, which contains 83k/41k/81k images for training/validation/test.
Following~\citet{ALBEF},
we use both training and validation splits for training, and include additional training samples from Visual Genome.
During inference on VQA,
we use the decoder to rank the 3,128 candidate answers~\cite{ALBEF,bilinear}.

\noindent\textbf{NLVR$^2$.}
We conduct experiment on the official split~\cite{NLVR}.

\noindent\textbf{VisDial.}
We finetune on the training split of VisDial v1.0 and evaluate on its validation set.

\begin{table}[!h]
	\centering	
	\begin{tabular}	{l   |  c  c  c }
		\toprule	 	
	 Task & init LR (ViT-L) & batch size & \#epoch \\
	 \midrule
	 Retrieval & $1e^{-5}$ ($5e^{-6}$) & 256 & 6  \\
	 Captioning & $1e^{-5}$ ($2e^{-6}$) & 256 & 5 \\
	 VQA & $2e^{-5}$ & 256 & 10 \\
	 NLVR$^2$ & $3e^{-5}$ & 256 & 15 \\
	 VisDial & $2e^{-5}$ & 240 & 20 \\ 		
		
		\bottomrule
	\end{tabular}
 	 \vspace{-1ex}	
	\caption
	{
	\small	
		Finetuning hyperparameters for downstream tasks.
	}
	\label{tbl:paramter}
 \vspace{-2ex}	
\end{table}

\begin{figure*}[!b]
\centering
  \includegraphics[width=\textwidth]{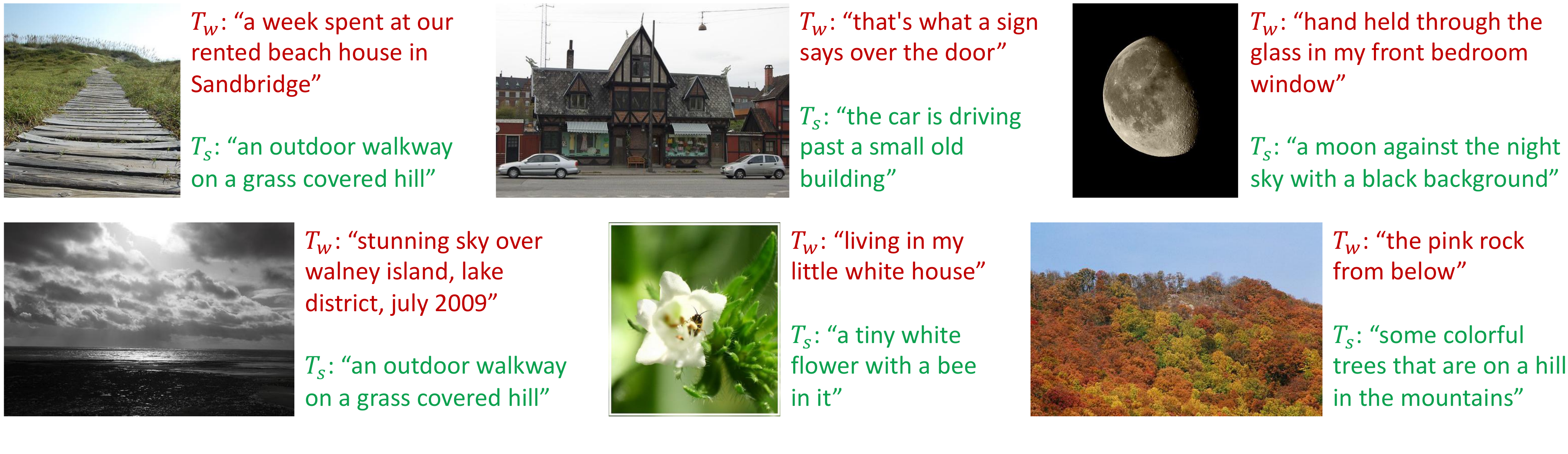}
\vspace{-6ex}
\caption{Examples of the web text $T_w$ and the synthetic text $T_s$. {\color{ForestGreen}Green} texts are accepted by the filter,
whereas {\color{BrickRed}red} texts are rejected.
}
\label{fig:more_example}
\end{figure*}

\vspace{-0.5ex}
\section{Additional Examples of Synthetic Captions}
In Figure~\ref{fig:more_example},
we show additional examples of images and texts where the web captions are filtered out, and the synthetic captions are kept as clean training samples.

\section{Pre-training Dataset Details}

Table~\ref{tbl:pretrain_data} shows the statistics of the pre-training datasets.
\begin{table}[!h]
	\small
	\centering	
	\setlength\tabcolsep{4pt}
	\resizebox{\columnwidth}{!}{%
		\begin{tabular}	{l   |  c c c c c c}
			\toprule
			  & COCO & VG  & SBU & CC3M & CC12M & LAION \\
			\midrule
			\# image  & 113K & 100K & 860K & 3M & 10M & 115M\\
			\# text & 567K & 769K & 860K & 3M & 10M & 115M\\
			\bottomrule
		\end{tabular}
	}
	\vspace{-2ex}	
	\caption
	{
		\small	
		Statistics of the pre-training datasets.
	}
	\label{tbl:pretrain_data}
	\vspace{-1ex}	
\end{table}

\end{document}